# Compiling Causal Theories to Successor State Axioms and STRIPS-Like Systems

**Fangzhen Lin** FLIN@CS.UST.HK
*Department of Computer Science*
*The Hong Kong University of Science and Technology*
*Clear Water Bay, Kowloon, Hong Kong*

## Abstract

We describe a system for specifying the effects of actions. Unlike those commonly used in AI planning, our system uses an action description language that allows one to specify the effects of actions using domain rules, which are state constraints that can entail new action effects from old ones. Declaratively, an action domain in our language corresponds to a nonmonotonic causal theory in the situation calculus. Procedurally, such an action domain is compiled into a set of logical theories, one for each action in the domain, from which fully instantiated successor state-like axioms and STRIPS-like systems are then generated. We expect the system to be a useful tool for knowledge engineers writing action specifications for classical AI planning systems, GOLOG systems, and other systems where formal specifications of actions are needed.

## 1. Introduction

We describe a system for generating action effect specifications from a set of domain rules and direct action effect axioms, among other things. We expect the system to be a useful tool for knowledge engineers writing action specifications for classical AI planning systems, GOLOG systems (Levesque et al., 1997), and other systems where formal specifications of actions are needed.

To motivate, consider the language used by STRIPS (Fikes & Nilsson, 1971) for describing the effects of actions. Briefly speaking, an action is described in this language by a first-order formula, called its precondition that describes the condition under which the action is executable, an add list that enumerates the propositions that the action will make true when successfully executed in a situation, and a delete list that enumerates the propositions that the action will make false when successfully executed in a situation. While the original STRIPS allowed the precondition and the elements of the two lists to be complex formulas, STRIPS actions now refer only to those whose precondition is given by a conjunction of atomic formulas and whose add and delete lists are lists of atomic formulas. It is widely acknowledged that this language is inadequate for describing actions in the real world. One of the limitations, the one that we address in this paper, is that with the language, one has to enumerate all possible effects of an action, a difficult if not impossible task for complex domains. For example, given a large C program, it is hard to figure out the effects of changing the value of a pointer on the values of all other pointers in the program. However, the underlying principle is very simple: when the value of a pointer changes, the values of all other pointers that point to the same memory location change as well. Put another way, the direct effect of the action of changing the value of a pointer to $x$ is that the value of the pointer will be $x$. The indirect or side effects of this action are those derived from the constraint which says that if two pointers point to a common location, then their values must be the same.

This idea of specifying the effects of actions using domain constraints is like "engineering from first principle", and has many advantages. First of all, constraints are action independent, and work on all actions. Secondly, if the effects of actions derived from domain constraints agree with one's expectation, then this will be a good indication that one has axiomatized the domain correctly. Finally, domain constraints can be used for other purposes as well. For instance, they can be used to check the consistency of the initial situation database. In general, when a set of sentences violates a domain constraint, we know that no legal situation can satisfy this set of sentences. This idea can and has been used in planning to prune impossible states. Recently, there have even been efforts at "reverse engineering" domain constraints from STRIPS-like systems to speed up planners (e.g. Zhang & Foo, 1997; Gerevini & Schubert, 1998; Fox & Long, 1998, and others).

While it is appealing to use domain constraints to derive the indirect effects of actions, making the idea work formally turned out to be a challenge. The problem is commonly known as the ramification problem, and various proposals have been made to solve it. Until recently, however, these proposals were at best of theoretical interest only because of their high computational complexity. The situation has since changed substantially due to the use of causality in representing domain constraints (Lin, 1995, 1996; McCain & Turner, 1995, 1997; Thielscher, 1995, 1997; Baral, 1995; Lifschitz, 1997, and others). What we will describe in this paper is an implemented system that builds on this recent work on causality-based approaches to the ramification problem. Specifically, our system takes as input an action domain description where actions are described by their precondition axioms and direct effect axioms, and domain constraints are represented by what we call *domain rules*. The system returns as output a complete action specification both in STRIPS-like format and as a set of fully instantiated successor state axioms (Reiter, 1991).

This paper is organized as follows. We begin by introducing our action description language. We then propose a procedure to compile an action domain specified in this language into a complete set of successor state axioms from which a STRIPS-like description is then extracted. We then show the soundness of this procedure with respect to a translation from action domain descriptions to situation calculus causal theories of Lin (1995). We next describe an implementation of this procedure, and present some experimental results. As one will see, one of the limitations of our system is that it is essentially propositional. While effect axioms and domain rules can have variables, they need to be fully instantiated during the compilation process. To partially overcome this limitation, we show some results that allow one to generalize the propositional output to the first-order case for certain classes of action domain descriptions. We then discuss some related work, and conclude this paper with some pointers for future work.

## 2. An Action Description Language

We assume a first-order language with equality. We shall call those predicates whose extensions may be changed by actions *fluents*, and those whose extensions are not changed by any actions *static relations*. We also call unary static relations *types*. By *fluent atoms* we mean those atomic formulas formed by fluents. An *equality* atom is one of the form $u = v$, where $u$ and $v$ are variables or constants, and an inequality constraint is one of the form $u \neq v$. Actions are represented by functions, and they are assumed to be the only functions with positive arities in the language.

Our action description language includes the following components.

### 2.1 Type Definitions

A type definition is specified by expressions of the following form:

$$Domain(p, \{a_1, ..., a_n\}),$$

where $p$ is a type, and $a_1, ..., a_n$ are constants. The intuitive meaning of this expression is that the domain (extension) of the type $p$ is the set $\{a_1, ..., a_n\}$. For instance, in the blocks world, we may have a type called *block*, and have, say, five blocks named numerically: $Domain(block, \{1, 2, 3, 4, 5\})$. In a logistics domain, we may have a type called *loc* for locations, and have, say, 3 locations $l_1$, $l_2$, and $l_3$: $Domain(loc, \{l_1, l_2, l_3\})$.

### 2.2 Primitive Fluent Definitions

Primitive fluents are defined by expressions of the following form:

$$Fluent(f(x_1, ..., x_n), p_1(x_1) \wedge \cdots \wedge p_n(x_n) \wedge e_1 \wedge \cdots \wedge e_m),$$

where $f$ is an n-ary fluent, each $p_i$, $1 \leq i \leq n$, a type, and each $e_i$, $1 \leq i \leq m$, an inequality constraint of the form $x_j \neq x_k$, for some $1 \leq j < k \leq n$. The intuitive meaning of this expression is that $f(x_1, ..., x_n)$ is a legal fluent atom if the second argument is true. For instance, in the blocks world, given the type definition $Domain(block, \{1, 2, 3\})$, the following fluent specification:

$$Fluent(on(x, y), block(x) \wedge block(y) \wedge x \neq y)$$

would generate the following six legal fluent atoms:

$$on(1, 2), on(1, 3), on(2, 1), on(2, 3), on(3, 1), on(3, 2).$$

Clearly, there should be exactly one fluent definition for each fluent.

### 2.3 Complex Fluent Definitions

Given a set of primitive fluents, one may want to define some new ones. For instance, in the blocks world, given the primitive fluent *on*, we can define *clear* in terms of *on* as: $(\forall x)clear(x) \equiv \neg(\exists y)on(y, x)$.

To specify complex fluents like *clear*, we first define *fluent formulas* as follows:

- $t_1 = t_2$ is a fluent formula, where $t_1$ and $t_2$ are terms, i.e. either a constant in the domain of a type or a variable.
- $f(t_1, ..., t_n)$ is a fluent formula, where $t_1, ..., t_n$ are terms, and $f$ is either an n-ary primitive fluent, a complex fluent, or a static relation.

- If $\varphi$ and $\varphi'$ are fluent formula, then $\neg\varphi$, $\varphi \vee \varphi'$, $\varphi \wedge \varphi'$, $\varphi \supset \varphi'$, and $\varphi \equiv \varphi'$ are also fluent formulas.
- If $\varphi$ is a fluent formula, $x$ a variable, and $p$ a type, then $\forall(x,p)\varphi$ (for all $x$ of type $p$, $\varphi$ holds) and $\exists(x,p)\varphi$ (for some $x$ of type $p$, $\varphi$ holds) are fluent formulas. Notice that we require types to have finite domains, so these quantifications are really shorthands: If the domain of $p$ is $\{a_1, ..., a_n\}$, then $\forall(x,p)\varphi$ stands for
$$\varphi(x/a_1) \wedge \cdots \wedge \varphi(x/a_n),$$
and $\exists(x,p)\varphi$ stands for
$$\varphi(x/a_1) \vee \cdots \vee \varphi(x/a_n).$$

A complex fluent is then specified in our language by a pair of expressions of the following form:

$$\begin{aligned}&\mathit{Complex}(f(x_1, ..., x_n), p_1(x_1) \wedge \cdots \wedge p_n(x_n) \wedge e_1 \cdots \wedge e_m),\\ &\mathit{Defined}(f(x_1, ..., x_n), \varphi),\end{aligned}$$

where $p_i$'s and $e_i$'s are the same as in primitive fluent definitions, and $\varphi$ a fluent formula that does not mention any complex fluents and whose free variables are among $x_1, ..., x_n$. The first expression specifies the syntax and the second the semantics of the complex fluent.

For instance, the complex fluent *clear* in the blocks world can be specified as:

$$\begin{aligned}&\mathit{Complex}(\mathit{clear}(x), \mathit{block}(x)),\\ &\mathit{Defined}(\mathit{clear}(x), \neg\exists(y, \mathit{block})\mathit{on}(y,x)).\end{aligned}$$

As we mentioned above, quantifiers here are just shorthands because each type must have a finite domain. For instance, given the following specification:

$$\begin{aligned}&\mathit{Domain}(\mathit{block}, \{1,2,3\}),\\ &\mathit{Fluent}(\mathit{on}(x,y), \mathit{block}(x) \wedge \mathit{block}(y))\end{aligned}$$

the above fluent definition for *clear* will be expanded to:

$$\begin{aligned}&\mathit{Defined}(\mathit{clear}(1), \neg(\mathit{on}(1,1) \vee \mathit{on}(2,1) \vee \mathit{on}(3,1))),\\ &\mathit{Defined}(\mathit{clear}(2), \neg(\mathit{on}(1,2) \vee \mathit{on}(2,2) \vee \mathit{on}(3,2))),\\ &\mathit{Defined}(\mathit{clear}(3), \neg(\mathit{on}(1,3) \vee \mathit{on}(2,3) \vee \mathit{on}(3,3))).\end{aligned}$$

### 2.4 Static Relation Definitions

As we mentioned, a static relation is one that is not changed by any action in the domain. For instance, in the robot navigation domain, we may have a proposition $\mathit{connected}(d, r1, r2)$ meaning that door $d$ connects rooms $r1$ and $r2$. The truth value of this proposition cannot be changed by the navigating robot that just rolls from room to room.

In our language, a static relation is defined by an expression of the following form:

$$\mathit{Static}(g(x_1, ..., x_n), p_1(x_1) \wedge \cdots \wedge p_n(x_n) \wedge e_1 \wedge \cdots \wedge e_m),$$

where $g$ is an n-ary predicate, and $p_i$'s and $e_i$'s are the same as in primitive fluent definitions. The meaning of this expression is similar to a fluent definition, and there should be exactly one definition for each static relation.

### 2.5 Domain Axioms

Domain axioms are constraints on static relations. For instance, for the static proposition $connected(d, r1, r2)$, we may want to impose the following constraint: $connected(d, r1, r2) \equiv connected(d, r2, r1)$. In our language, domain axioms are specified by expressions of the form:

$$Axiom(\varphi),$$

where $\varphi$ is a fluent formula that does not mention any fluents, i.e. it mentions only static relations and equality. For instance, the above constraint on *connected* is written as:

$$Axiom(\forall(d, door)\forall(r_1, room)\forall(r_2, room)connected(d, r_1, r_2) \equiv connected(d, r_2, r_1)),$$

where *door* and *room* are types.

### 2.6 Action Definitions

Actions are defined by expressions of the following form:

$$Action(a(x_1, ..., x_n), p_1(x_1) \wedge \cdots \wedge p_n(x_n) \wedge e_1 \wedge \cdots \wedge e_m),$$

where $a$ is an n-ary action, and $p_i$'s and $e_i$'s are the same as in primitive fluent definitions. For instance, in the blocks world, given the type definition $Domain(block, \{1, 2, 3\})$, the following action specification:

$$Action(stack(x, y), block(x) \wedge block(y) \wedge x \neq y)$$

would generate the following six action instances:

$$stack(1, 2), stack(1, 3), stack(2, 1), stack(2, 3), stack(3, 1), stack(3, 2).$$

There should be exactly one action definition for each action.

### 2.7 Action Precondition Definitions

Action precondition definitions are specified by expressions of the following form:

$$Precond(a(x_1, ..., x_n), \varphi),$$

where $a$ is an n-ary action, $\varphi$ is a fluent formula whose free variables are among $x_1, ..., x_n$.

There should be exactly one precondition definition for each action. For instance, in the blocks world, we may have:

$$Precond(stack(x, y), clear(x) \wedge clear(y) \wedge ontable(x)),$$

which says that for the action $stack(x, y)$ to be executable in a situation, $clear(x)$, $clear(y)$, and $ontable(x)$ must be true in it.

### 2.8 Action Effect Specifications

Action effects are specified by expressions of the following form:

$$Effect(a(x_1, ..., x_n), \varphi, f(y_1, ..., y_k)),$$

or of the form:

$$Effect(a(x_1, ..., x_n), \varphi, \neg f(y_1, ..., y_k)),$$

where $\varphi$ is a fluent formula, and $f$ a primitive fluent. The intuitive meaning of these expressions is that if $\varphi$ is true in the initial situation, then action $a(x_1, ..., x_n)$ will cause $f(y_1, ..., y_k)$ to be true (false). For instance, in the blocks world, action $stack(x, y)$ causes $x$ to be on $y$:

$$Effect(stack(x, y), true, on(x, y)).$$

For an example of a context dependent effect, consider action $drop(x)$ that breaks an object only if it is fragile:

$$Effect(drop(x), fragile(x), broken(x)).$$

Notice here that the fluent formula $\varphi$ in the action effect specifications can have variables that are not in $x_1, ..., x_n, y_1, ..., y_k$. Informally, all the variables are supposed to be "universally quantified." More precisely, when these expressions are instantiated, one can substitute any objects for these variables, provided the resulting formulas are well-formed. For instance, given the action effect specification $Effect(move(x), g(x_1) \wedge q(x_1, x_2), f(y))$, one can instantiate it to: $Effect(move(a), g(b) \wedge q(b, c), f(d))$, as long as $move(a)$ is a legal action (according to the action definition for *move*) and $g(b)$, $q(b, c)$, and $f(d)$ are legal fluent atoms (according to the fluent definitions for $g$, $q$, and $f$).

### 2.9 Domain Rules

Domain rules are specified by expressions of the following form:

$$Causes(\varphi, f(x_1, ..., x_n)),$$

or of the following form:

$$Causes(\varphi, \neg f(x_1, ..., x_n)),$$

where $\varphi$ is a fluent formula, and $f$ a primitive fluent. Like action effect specifications, $\varphi$ here can have variables that are not in $x_1, ..., x_n$. The intuitive meaning of a domain rule is that in any situation, if $\varphi$ holds, then the fluent atom $f(x_1, ..., x_n)$ will be caused to be true. A domain rule is stronger than material implication. Its formal semantics is given by mapping it to a causal rule of Lin (1995) (see Section 4), thus the name "causes". For instance, in the blocks world, a block can be on only one other block:

$$Causes(on(x, y) \wedge y \neq z, \neg on(x, z)).$$

In a logistics domain, one may want to say that if a package is inside a truck which is at location $l$, then the package is at location $l$ as well:

$$Causes(in(x, y) \wedge at(y, l), at(x, l)).$$

### 2.10 Action Domain Descriptions

The following definition sums up our action description language:

**Definition 1** *An* action domain description *is a set of type definitions, primitive fluent definitions, complex fluent definitions, static proposition definitions, domain axioms, action definitions, action precondition definitions, action effect specifications, and domain rules.*

**Example 1** The following action domain description defines a blocks world with three blocks:

$$\begin{aligned}
&Domain(block, \{1,2,3\}),\\
&Fluent(on(x,y), block(x) \wedge block(y)),\\
&Fluent(ontable(x), block(x)),\\
&Complex(clear(x), block(x)),\\
&Defined(clear(x), \neg\exists(y, block) on(y,x)),
\end{aligned}$$

$$\begin{aligned}
&Causes(on(x,y) \wedge x \neq z, \neg on(z,y)),\\
&Causes(on(x,y) \wedge y \neq z, \neg on(x,z)),\\
&Causes(on(x,y), \neg ontable(x)),\\
&Causes(ontable(x), \neg on(x,y)),
\end{aligned}$$

$$\begin{aligned}
&Action(stack(x,y), block(x) \wedge block(y) \wedge x \neq y),\\
&Precond(stack(x,y), ontable(x) \wedge clear(x) \wedge clear(y)),\\
&Effect(stack(x,y), true, on(x,y)),
\end{aligned}$$

$$\begin{aligned}
&Action(unstack(x,y), block(x) \wedge block(y) \wedge x \neq y),\\
&Precond(unstack(x,y), clear(x) \wedge on(x,y)),\\
&Effect(unstack(x,y), true, ontable(x)),
\end{aligned}$$

$$\begin{aligned}
&Action(move(x,y,z), block(x) \wedge block(y) \wedge block(z) \wedge x \neq y \wedge x \neq z \wedge y \neq z),\\
&Precond(move(x,y,z), on(x,y) \wedge clear(x) \wedge clear(z)),\\
&Effect(move(x,y,z), true, on(x,z)).
\end{aligned}$$

## 3. A Procedural Semantics

Given an action domain description $\mathcal{D}$, we use the following procedure called CCP (Causal Completion Procedure) to generate a complete action effect specification:

1. Use primitive and complex fluent definitions to generate all legal fluent atoms. In the following let $\mathcal{F}$ be the set of fluent atoms so generated.

2. Use action definitions to generate all legal action instances, and for each such action instance $A$ do the following.

2.1. For each primitive fluent atom $F \in \mathcal{F}$, collect all ground instances[1] of $A$'s positive effects:

$$Effect(A, \varphi_1, F), \cdots, Effect(A, \varphi_n, F),$$

all ground instances of $A$'s negative effects:

$$Effect(A, \phi_1, \neg F), \cdots, Effect(A, \phi_m, \neg F),$$

all ground instances of positive domain rules:

$$Causes(\varphi'_1, F), \cdots, Causes(\varphi'_k, F),$$

all ground instances of negative domain rules:

$$Causes(\phi'_1, \neg F), \cdots, Causes(\phi'_l, \neg F),$$

and generate the following pseudo successor state axiom;

$$\begin{aligned} succ(F) \equiv\ & init(\varphi_1) \vee \cdots \vee init(\varphi_n) \vee succ(\varphi'_1) \vee \cdots \vee succ(\varphi'_l) \vee \\ & init(F) \wedge \neg[init(\phi_1) \vee \cdots \vee init(\phi_m) \vee \\ & succ(\phi'_1) \vee \cdots \vee succ(\phi'_k)], \end{aligned} \quad (1)$$

where for any fluent formula $\varphi$, $init(\varphi)$ is the formula obtained from $\varphi$ by replacing every fluent atom $f$ in $\varphi$ by $init(f)$, and similarly $succ(\varphi)$ is the formula obtained from $\varphi$ by replacing every fluent atom $f$ in $\varphi$ by $succ(f)$. Intuitively, $init(f)$ means that $f$ is true in the initial situation, and $succ(f)$ that $f$ is true in the successor situation of performing action $A$ in the initial situation.

2.2. Let $Succ$ be the set of pseudo successor state axioms, one for each primitive fluent $F$, generated by the last step, $Succ1$ the following set of axioms:

$$Succ1 = \{succ(F) \equiv succ(\varphi) \mid Defined(F, \varphi) \text{ is a complex fluent definition}\},$$

and $Init$ the following set of axioms:

$$\begin{aligned} Init = \{ & \varphi \mid Axiom(\varphi) \text{ is a domain axiom}\} \cup \\ & \{init(\varphi) \supset init(F) \mid Causes(\varphi, F) \text{ is a domain rule}\} \cup \\ & \{init(\varphi) \supset \neg init(F) \mid Causes(\varphi, \neg F) \text{ is a domain rule}\} \cup \\ & \{init(F) \equiv init(\varphi) \mid Defined(F, \varphi) \text{ is a complex fluent definition}\} \cup \\ & \{init(\phi_A) \mid Precond(A, \phi_A) \text{ is the precondition definition for } A\}. \end{aligned}$$

For each fluent atom $F$, if there is a formula $\Phi_F$ such that

$$Init \cup Succ \cup Succ1 \models succ(F) \equiv \Phi_F,$$

and $\Phi_F$ does not mention propositions of the form $succ(f)$, then output the axiom

$$succ(F) \equiv \Phi_F.$$

1. When generating ground instances, all shorthands like $\forall(x, p)$ are expanded. See the definition of fluent formulas in the last section.

Otherwise, the action $A$'s effect on $F$ is indeterminate. In this case, output two axioms:

$$succ(F) \supset \alpha_F,$$
$$\beta_F \supset succ(F),$$

where $\alpha_F$ is a strongest formula satisfying the first implication, and $\beta_F$ a weakest formula satisfying the second implication. In the following, to be explicit about the action $A$ for which we are computing its effects, we will write the axioms as $Succ_A$, $Init_A$, and $Succ1_A$.

Conceptually, Step 2.1 in the above procedure is most significant. In the next section, we shall prove that this step is provably correct under a translation to the situation calculus causal theories of Lin (1995). Computationally, Step 2.2 is the most expensive. We shall describe the strategies that our system uses to implement it in Section 5.

For this procedure to work properly, the action domain description should satisfy the following conditions.

1. We require that all fluent atoms in $Init$, $Succ$, and $Succ1$ be among those generated in Step 1. This would rule out cases like

   $$Fluent(on(x,y), block(x) \wedge block(y) \wedge x \neq y)$$

   together with $Defined(clear(x), \neg\exists(y, block)on(y,x))$, as the latter would generate fluent atoms of the form $on(x,x)$ which are ruled out by the fluent definition for *on*. Here one could either drop the inequality constraint in the definition of *on* or change the complex fluent definition into $Defined(clear(x), \neg\exists(y, block)(on(y,x) \wedge x \neq y))$. We could have built in a test in our procedure above to reject an action domain description with incoherent fluent definitions like this. One easy way of making sure this does not happen is not to use inequality constraints in the definition of fluents.

2. As we mentioned above, for each action there should be exactly one action precondition that captures exactly the conditions under which the action is executable. When the action precondition is given explicitly like this, one needs to be careful in writing action effect axioms and domain rules so that no contradictory effects would be generated. For instance, given $Precond(A, true)$, the action effect axioms $Effect(A, true, F)$ and $Effect(A, true, \neg F)$ are clearly not realizable simultaneously. Similarly, if $Causes(true, F)$ is given as a domain rule, then one should not write the effect axiom $Effect(A, true, \neg F)$. Had we not insisted that $A$ be always executable, we could simply conclude that $A$ is not executable when its effect axioms are in contradiction or when some of its effect axioms contradict domain rules. It remains future work to extend our procedure to allow for automatic generation of these implicitly given action preconditions. For now, we shall assume that the given action domain specification is consistent in the sense that for each action instance $A$ generated in Step 1, the following theory

   $$Init \cup Succ \cup Succ1 \cup \{init(\varphi) \supset \neg succ(F) \mid Effect(A, \varphi, \neg F) \text{ is a ground instance of effect axiom}\}$$

is consistent.

3. On a related point, our procedure assumes that information about the initial situation is given by $Init$. In particular, action effect axioms should not entail any information about the initial situation. For instance, given $Causes(q, \neg p)$, $Effect(A, true, p)$, and $Precond(A, true)$, it must be that in the initial situation, $q$ cannot be true, for otherwise, it will persist into the next situation, causing $p$ to be false, which contradicts the action effect. Formally, this means that given any set $I$ of atoms of the form $init(f)$, where $f$ is a primitive fluent atom, if $I \cup I^- \cup Init$ is consistent, then $I \cup I^- \cup Succ \cup Succ1$ is also consistent, where $I^-$, the complement of $I$, is the following set:

$$\{\neg init(f) \mid init(f) \notin I \text{ and } f \text{ is a primitive fluent atom generated by Step 1}\}$$

   Notice that for a similar reason, Reiter needed what he called the consistency assumption in order for his completion procedure to be sound and complete for generating successor state axioms (Reiter, 1991).

While our action domain descriptions are clearly targeted at specifying deterministic actions, some indeterminate effects can sometimes arise from cyclic domain rules. For instance, consider the following action domain description:

$$Causes(p, p),$$
$$Precond(A, true)$$

For action $A$, $Init$ is a tautology, $Succ1$ is empty, and $Succ$ consists of the following pseudo-successor state axiom for $p$:

$$succ(p) \equiv succ(p) \vee init(p),$$

which is equivalent to $init(p) \supset succ(p)$. So if initially $p$ is true, then after $A$ is performed, we know that $p$ will continue to be true. But if $p$ is initially false, then after $A$ is performed, we do not know if $p$ is true or not.

**Example 2** Consider the blocks world description in Example 1. The set of fluent atoms generated by Step 1 is:

$$\mathcal{F} = \{on(1,1), on(1,2), on(1,3), on(2,1), on(2,2), on(2,3), on(3,1), on(3,2), on(3,3), clear(1), clear(2), clear(3), ontable(1), ontable(2), ontable(3)\}.$$

Step 2 generates the following action instances:

$$stack(1,2), stack(1,3), stack(2,1), stack(2,3), stack(3,1), stack(3,2), unstack(1,2), unstack(1,3), unstack(2,1), unstack(2,3), unstack(3,1), unstack(3,2), move(1,2,3), move(1,3,2), move(2,1,3), move(2,3,1), move(3,1,2), move(3,2,1)$$

For each of these action instances, we need to go through Steps 2.1 and 2.2. For instance, for $stack(1,2)$, there is only one effect axiom about $on(1,2)$:

$$Effect(stack(1,2), true, on(1,2)),$$

and the following causal rules about $on(1,2)$:

$$\begin{array}{l} Causes(on(1,1), \neg on(1,2)), \\ Causes(on(1,3), \neg on(1,2)), \\ Causes(on(2,2), \neg on(1,2)), \\ Causes(on(3,2), \neg on(1,2)), \\ Causes(ontable(1), \neg on(1,2)). \end{array}$$

Therefore Step 2.1 generates the following pseudo-successor state axiom for $on(1,2)$:

$$\begin{array}{l} succ(on(1,2)) \equiv true \lor \\ \quad init(on(1,2)) \land \neg[succ(on(1,1)) \lor succ(on(1,3)) \lor \\ \quad\quad succ(on(2,2)) \lor succ(on(3,2) \lor succ(ontable(1))]. \end{array}$$

We can similarly generate the following pseudo-successor state axioms for the other primitive fluent atoms:

$$\begin{array}{l} succ(on(1,1)) \equiv init(on(1,1)) \land \neg[succ(on(1,2)) \lor succ(on(1,3)) \lor \\ \quad\quad succ(on(2,1)) \lor succ(on(3,1) \lor succ(ontable(1))], \\ succ(on(1,3)) \equiv init(on(1,3)) \land \neg[succ(on(1,1)) \lor succ(on(1,2)) \lor \\ \quad\quad succ(on(2,3)) \lor succ(on(3,3)) \lor succ(ontable(1))], \\ succ(on(2,1)) \equiv init(on(2,1)) \land \neg[succ(on(1,1)) \lor succ(on(3,1)) \lor \\ \quad\quad succ(on(2,2)) \lor succ(on(2,3)) \lor succ(ontable(2))], \\ succ(on(2,2)) \equiv init(on(2,2)) \land \neg[succ(on(1,2)) \lor succ(on(1,3)) \lor \\ \quad\quad succ(on(2,1) \lor succ(on(2,3)) \lor succ(ontable(2))], \\ succ(on(2,3)) \equiv init(on(2,3)) \land \neg[succ(on(1,3)) \lor succ(on(3,3)) \lor \\ \quad\quad succ(on(2,1) \lor succ(on(2,2)) \lor succ(ontable(2))], \\ succ(on(3,1)) \equiv init(on(3,1)) \land \neg[succ(on(3,2)) \lor succ(on(3,3)) \lor \\ \quad\quad succ(on(1,1) \lor succ(on(1,3)) \lor succ(ontable(1))], \\ succ(on(3,2)) \equiv init(on(3,2)) \land \neg[succ(on(3,1)) \lor succ(on(3,3)) \lor \\ \quad\quad succ(on(1,2) \lor succ(on(2,2)) \lor succ(ontable(2))], \\ succ(on(3,3)) \equiv init(on(3,3)) \land \neg[succ(on(3,1)) \lor succ(on(3,2)) \lor \\ \quad\quad succ(on(1,3) \lor succ(on(2,3)) \lor succ(ontable(3))], \\ succ(ontable(1)) \equiv \\ \quad init(ontable(1)) \land \neg[succ(on(1,2)) \lor succ(on(1,1)) \lor succ(on(1,3))], \\ succ(ontable(2)) \equiv \\ \quad init(ontable(2)) \land \neg[succ(on(2,1)) \lor succ(on(2,2)) \lor succ(on(2,3))], \end{array}$$

$$succ(ontable(3)) \equiv$$
$$init(ontable(3)) \wedge \neg[succ(on(3,1)) \vee succ(on(3,2)) \vee succ(on(3,3))].$$

For the complex fluent *clear*, its definition yields the following axioms:

$$succ(clear(1)) \equiv \neg succ(on(1,1)) \wedge \neg succ(on(2,1)) \wedge \neg succ(on(3,1)),$$
$$succ(clear(2)) \equiv \neg succ(on(1,2)) \wedge \neg succ(on(2,2)) \wedge \neg succ(on(3,2)),$$
$$succ(clear(3)) \equiv \neg succ(on(1,3)) \wedge \neg succ(on(2,3)) \wedge \neg succ(on(3,3)).$$

We can then "solve" these pseudo-successor state axioms and generate the following successor state axioms:

| | |
|---|---|
| $succ(on(1,1)) \equiv false$ | $succ(on(1,2)) \equiv true$ |
| $succ(on(1,3)) \equiv false$ | $succ(on(2,1)) \equiv false$ |
| $succ(on(2,2)) \equiv false$ | $succ(on(2,3)) \equiv init(on(2,3))$ |
| $succ(on(3,1)) \equiv false$ | $succ(on(3,2)) \equiv false$ |
| $succ(on(3,3)) \equiv init(on(3,3))$ | $succ(ontable(1)) \equiv false$ |
| $succ(ontable(2)) \equiv init(ontable(2))$ | $succ(ontable(3)) \equiv init(ontable(3))$ |
| $succ(clear(1)) \equiv init(clear(1))$ | $succ(clear(2)) \equiv false$ |
| $succ(clear(3)) \equiv init(clear(3))$ | |

Once we have a set of these fully instantiated successor state axioms, we then generate STRIPS-like descriptions like the following:

```
stack(1, 2)
Preconditions:  ontable(1), clear(1), clear(2).
Add list:       on(1, 2).
Delete list:    ontable(1), clear(2).
Cond. effects:  none.
Indet. effects: none.

stack(1, 3)
Preconditions:  ontable(1), clear(1), clear(3).
Add list:       on(1,3).
Delete list:    ontable(1), clear(3).
Cond. effects:  none.
Indet. effects: none.
```

We have the following remarks:

- Although we generate the axiom $succ(on(1,3)) \equiv false$ for $stack(1,2)$, we do not put $on(1,3)$ into its delete list. This is because we can deduce $init(on(1,3)) \equiv false$ from $Init$ as well. A fluent atom is put into the add or the delete list of an action only if this fluent atom's truth value is definitely changed by the action. See Section 5 for more details about how a STRIPS-like description is generated from successor state axioms.
- As one can see, our CCP procedure crucially depends on the fact that each type has a finite domain so that all reasoning can be done in propositional logic. This is a limitation of our current system, and this limitation is not as bad as one might think. First of all, typical planning problems all assume finite domains, and changing the domain of a type in an action description is easy - all one needs to do is to change the corresponding type definition. More significantly, a generic action domain description can often be obtained from one that assumes a finite domain. In our blocks world example, the numbers "1", "2", and "3" are generic names, and can be replaced by parameters. For instance, if we replace "1" by $x$ and "2" by $y$ in the above STRIPS-like description of $stack(1,2)$, we will get a STRIPS-like description for $stack(x,y)$ that works for any $x$ and $y$. We have found that this is a strategy that often works in planning domains.

## 4. Formal Semantics

The formal semantics of an action domain description is defined by translating it into a situation calculus causal theory of Lin (1995). We shall show that the procedure CCP given above is sound under this semantics.

This section is mainly for those who are interested in nonmonotonic action theories. For those who are interested only in using our action description language for describing action domains, this section can be safely skipped.

We first briefly review the language of the situation calculus.

### 4.1 Situation Calculus

The language of the situation calculus is a many sorted first-order language. We assume the following sorts: *situation* for situations, *action* for actions, *fluent* for propositional fluents, *truth-value* for truth values $true$ and $false$, and *object* for everything else.

We use the following domain independent predicates and functions:

- Binary function $do$ - for any action $a$ and any situation $s$, $do(a,s)$ is the situation resulting from performing $a$ in $s$.
- Binary predicate $H$ - for any $p$ and any situation $s$, $H(p,s)$ is true if $p$ holds in $s$.
- Binary predicate $Poss$ - for any action $a$ and any situation $s$, $Poss(a,s)$ is true if $a$ is possible (executable) in $s$.
- Ternary predicate $Caused$ - for any fluent atom $p$, any truth value $v$, and any situation $s$, $Caused(p,v,s)$ is true if the fluent atom $p$ is caused (by something unspecified) to have the truth value $v$ in the situation $s$.

In the last section, we introduced fluent formulas. We now extend $H$ to these formulas: for any fluent formula $\varphi$ and situation $s$, $H(\varphi, s)$ is defined as follows:

- $H(t_1 = t_2, s)$ is $t_1 = t_2$.
- if $P$ is a static proposition, then $H(P, s)$ is $P$.
- inductively, $H(\neg\varphi, s)$ is $\neg H(\varphi, s)$, $H(\varphi \vee \varphi', s)$ is $H(\varphi, s) \vee H(\varphi', s)$, and similarly for other connectives.
- inductively, $H(\forall(x,p)\varphi, s)$ is $\forall x.[p(x) \supset H(\varphi, s)]$ and $H(\exists(x,p)\varphi, s)$ is $\exists x.[p(x) \wedge H(\varphi, s)]$.

According to this definition, $H(\varphi, s)$ will be expanded to a situation calculus formula where $H$ is applied to fluents.

### 4.2 A Translation to the Situation Calculus

Given a first-order language $\mathcal{L}$ for writing action domain descriptions, we assume that there will be a corresponding language $\mathcal{L}'$ for the situation calculus such that constants in $\mathcal{L}$ will be constants of sort *object* in $\mathcal{L}'$, types in $\mathcal{L}$ will be types (unary predicates) in $\mathcal{L}'$, static relations will be predicates of the same arities in $\mathcal{L}'$, fluents in $\mathcal{L}$ will be functions of sort *fluent* in $\mathcal{L}'$, and actions in $\mathcal{L}$ will be functions of sort *action* in $\mathcal{L}'$. Under these conventions, the following translation will map an action domain description to a situation calculus theory.

Let $\mathcal{D}$ be an action domain description. The translation of $\mathcal{D}$ into a situation calculus theory is defined as follows:

- a type definition $Domain(p, \{a_1, ..., a_k\})$ is translated to:

$$(\forall x).p(x) \equiv (x = a_1 \vee \cdots \vee x = a_k),$$
$$a_1 \neq a_2 \neq \cdots \neq a_k.$$

- a primitive fluent definition

$$Fluent(f(x_1, ..., x_n), p_1(x_1) \wedge \cdots \wedge p_n(x_n) \wedge e_1 \wedge \cdots \wedge e_m)$$

  is translated to

$$(\forall x_1, ..., x_n).Fluent(f(x_1, ..., x_n)) \equiv p_1(x_1) \wedge \cdots \wedge p_n(x_n) \wedge e_1 \wedge \cdots \wedge e_m.$$

- a complex fluent definition

$$Complex(f(x_1, ..., x_n), p_1(x_1) \wedge \cdots \wedge p_n(x) \wedge e_1 \cdots \wedge e_m),$$
$$Defined(f(x_1, ..., x_n), \varphi),$$

  is translated to

$$(\forall x_1, ..., x_n).Fluent(f(x_1, ..., x_n)) \equiv p_1(x_1) \wedge \cdots \wedge p_n(x) \wedge e_1 \cdots \wedge e_m,$$
$$(\forall x_1, ..., x_n, s).Fluent(f(x_1, ..., x_n)) \supset [H(f(x_1, ..., x_n), s) \equiv H(\varphi, s)].$$

- a domain axiom about static propositions:

$$Axiom(\varphi)$$

is translated to $\varphi$ with quantifiers in it treated as shorthands:

$$\forall(x,p)\phi = \forall x.p(x) \supset \phi,$$
$$\exists(x,p)\phi = \exists x.p(x) \wedge \phi.$$

- an action definition

$$Action(a(x_1,...,x_n), p_1(x_1) \wedge \cdots \wedge p_n(x_n) \wedge e_1 \wedge \cdots \wedge e_m)$$

is translated to

$$(\forall x_1,...,x_n).Action(a(x_1,...,x_n)) \equiv p_1(x_1) \wedge \cdots \wedge p_n(x_n) \wedge e_1 \wedge \cdots \wedge e_m.$$

We assume here that the domain description has only one action definition for each action.

- An action precondition axiom

$$Precond(a(x_1,...,x_n), \varphi)$$

is translated to

$$(\forall \vec{\xi}, s).Action(a(x_1,...,x_n)) \supset [Poss(a(x_1,...,x_n), s) \equiv H(\varphi, s)],$$

where $\vec{\xi}$ is the list of all free variables in $a(x_1,...,x_n)$ and $\varphi$. We mentioned earlier that one of the limitations of our current system is that action preconditions have to be given explicitly. This is reflected in the above translation.

- An action effect axiom:

$$Effect(a(x_1,...,x_n), \varphi, f(y_1,...,y_k)),$$

is translated to

$$(\forall \vec{\xi}, s).Action(a(x_1,...,x_n)) \wedge Fluent(f(y_1,...,y_k)) \wedge Poss(a(x_1,...,x_n), s) \supset$$
$$\{H(\varphi, s) \supset Caused(f(y_1,...,y_k), true, do(a(x_1,...,x_n), s))\},$$

where $\vec{\xi}$ is the list of all free variables in $a(x_1,...,x_n)$, $f(y_1,...,y_k)$, and $\varphi$.

Similarly, an effect axiom

$$Effect(a(x_1,...,x_n), \varphi, \neg f(y_1,...,y_k)),$$

is translated to

$$(\forall \vec{\xi}, s).Action(a(x_1,...,x_n)) \wedge Fluent(f(y_1,...,y_k)) \wedge Poss(a(x_1,...,x_n), s) \supset$$
$$\{H(\varphi, s) \supset Caused(f(y_1,...,y_k), false, do(a(x_1,...,x_n), s))\}.$$

- A domain rule of the form

  $$Causes(\varphi, f(x_1, ..., x_n))$$

  is translated to

  $$(\forall\vec{\xi}).Fluent(f(x_1, ..., x_n)) \supset (\forall s)\{H(\varphi, s) \supset Caused(f(x_1, ..., x_n), true, s)\},$$

  where $\vec{\xi}$ is the list of all free variables in $f(x_1, ..., x_n)$ and $\varphi$. Similarly, a domain rule of the form

  $$Causes(\varphi, \neg f(x_1, ..., x_n))$$

  is translated to

  $$(\forall\vec{\xi}).Fluent(f(x_1, ..., x_n)) \supset (\forall s)\{H(\varphi, s) \supset Caused(f(x_1, ..., x_n), false, s)\}.$$

Now given an action domain description $\mathcal{D}$, let $\mathcal{T}$ be its translation in the situation calculus. The semantics of $\mathcal{T}$ is then determined by its completion $comp(\mathcal{T})$ that is defined as the set of following sentences:

1. The circumscription of *Caused* in $T$ with all other predicates fixed.

2. The following basic axioms about *Caused* that says that if a fluent atom is caused to be true (false), then it is true (false):

   $$Caused(p, true, s) \supset Holds(p, s), \tag{2}$$

   $$Caused(p, false, s) \supset \neg Holds(p, s). \tag{3}$$

3. For the truth values, the following unique names and domain closure axiom:

   $$true \neq false \wedge (\forall v)(v = true \vee v = false). \tag{4}$$

4. The unique names assumptions for fluents and actions. Specifically, if $F_1$,..., $F_n$ are all the fluents, then we have:

   $$F_i(\vec{x}) \neq F_j(\vec{y}), \; i \text{ and } j \text{ are different},$$

   $$F_i(\vec{x}) = F_i(\vec{y}) \supset \vec{x} = \vec{y}.$$

   Similarly for actions.

5. For each primitive fluent atom $F$, the following generic successor state axiom:

   $$\forall a, s.Poss(a, s) \supset H(F, do(a, s)) \equiv \\ [Caused(F, true, do(a, s)) \vee H(F, s) \wedge \neg Caused(F, false, do(a, s))]. \tag{5}$$

6. The *foundational axioms* (Lin & Reiter, 1994b) for the *discrete* situation calculus. These axioms characterize the structure of the space of situations. For the purpose of this paper, it is enough to mention that they include the following unique names axioms for situations:

   $$s \neq do(a, s),$$

   $$do(a, s) = do(a', s') \supset (a = a' \wedge s = s').$$

The following theorem shows that the procedural semantics given in the previous section is sound with respect to the semantics given here.

**Theorem 1** *Let $\mathcal{D}$ be an action domain description, and $\mathcal{T}$ its translation in the situation calculus. Let $A$ be any ground action instance, $s$ a situation variable, and $\varphi(s)$ a situation calculus formula that satisfies the following two conditions (1) it contains at most the two situation terms $s$ and $do(A, s)$; and (2) it does not mention any predicate other than $H$, equality, and static relations. Let $\hat{\varphi}$ be obtained from $\varphi$ by replacing each $H(f, s)$ in it by $init(f)$, and each $H(f, do(A, s))$ in it by $succ(f)$. Then*

$$comp(\mathcal{T}) \models \forall s.Poss(A, s) \supset \varphi(s) \tag{6}$$

*if*

$$Init \cup Succ \cup Succ_1 \models \hat{\varphi}, \tag{7}$$

*where $Init$, $Succ$, and $Succ_1$ are the sets of axioms generated for $A$ according to the procedure in Section 3*

**Proof:** Suppose $S$ is a situation and $M$ is a model of $comp(\mathcal{T}) \cup \{Poss(A, S)\}$. Construct $M_{S,A}$ as follows:

- the domain of $M_{S,A}$ is the object domain of $M$.
- the interpretations of non-situational function and predicate symbols in $M_{S,A}$ are the same as those in $M$.
- for any fluent atom $f$, $M_{S,A} \models init(f)$ iff $M \models H(f, S)$ and $M_{S,A} \models succ(f)$ iff $M \models H(f, do(A, S))$.

Clearly, $M \models \varphi(S)$ iff $M_{S,A} \models \hat{\varphi}$. We show below that $M_{S,A}$ is a model of the left hand side of (7). From this, we see that if (7), then (6).

Notice first that a ground fluent atom $F$ is generated by the procedure CCP iff $Fluent(F)$ is true in $M$. Notice also that all fluent atoms in $Init \cup Succ \cup Succ_1$ must be generated by the procedure.

We show first that $M_{S,A}$ is a model of $Init$:

1. If $Axiom(\psi)$ is a domain axiom, then $\psi$ is in $\mathcal{T}$. Thus $M$ satisfies $\psi$. Since $\psi$ has no fluent symbols in it, $M_{S,A}$ satisfies it too.
2. If $Causes(\psi, f(x_1, ..., x_n))$ is a domain rule, then

$$(\forall \vec{\xi}).Fluent(f(x_1, ..., x_n)) \supset (\forall s)\{H(\psi, s) \supset Caused(f(x_1, ..., x_n), true, s)\}$$

is in $\mathcal{T}$. Thus $M$ satisfies

$$(\forall \vec{\xi}).Fluent(f(x_1, ..., x_n)) \supset (\forall s)\{H(\psi, s) \supset H(f(x_1, ..., x_n), s)\}.$$

Thus if $init(\psi') \supset init(F)$ is a corresponding formula in $Init$, where $\psi'$ and $F$ is some ground instantiation of $\psi$ and $f(x_1, ..., x_n)$, respectively, then $Fluent(F)$ must be true in $M$ (otherwise the above formula would not be in $Init$), so $M$ satisfies $H(\psi', S) \supset H(F, S)$. By the construction of $M_{S,A}$, it satisfies $init(\psi') \supset init(F)$. The case for $Causes(\psi, \neg f)$ is similar.

3. Suppose $\mathit{Defined}(F, \psi)$ is an instantiation of a complex fluent definition such that $F \equiv \psi$ is in $\mathit{Init}$. For it to be in $\mathit{Init}$, $\mathit{Fluent}(F)$ must be true. Thus $M$ must satisfy $H(F, S) \equiv H(\psi, S)$. By the construction of $M_{S,A}$, it satisfies $\mathit{init}(F) \equiv \mathit{init}(\psi, S)$.

4. Suppose $\mathit{Precond}(A, \phi_A)$ is the precondition axiom for $A$. Since $M$ satisfies $\mathit{Poss}(A, S)$ and $\mathit{Action}(A)$ (because $A$ is one of the action instances generated by the procedure), thus $M$ satisfies $H(\phi_A, S)$. So $M_{S,A}$ satisfies $\mathit{init}(\phi_A)$.

We now show that $M_{S,A}$ is a model of $\mathit{Succ}$, that is, for each primitive fluent atom $F$ generated by the procedure in Step 1, the pseudo-successor state axiom (1) holds. Referring to the notation in the axiom, we need to show that $M$ satisfies the following formula:

$$\begin{aligned} H(F, do(A, S)) \equiv\ & H(\varphi_1, S) \vee \cdots \vee H(\varphi_n, S) \vee \\ & H(\varphi'_1, do(A, S)) \vee \cdots \vee H(\varphi'_l, do(A, S)) \vee \\ & H(F, S) \wedge \neg[H(\phi_1, S) \vee \cdots \vee H(\phi_m, S) \vee \\ & \qquad H(\phi'_1, do(A, S)) \vee \cdots \vee H(\phi'_k, do(A, S))]. \end{aligned}$$

First of all, instantiating the generic successor state axiom (5) over $A$ and $S$, we get:

$$\begin{aligned} & \mathit{Poss}(A, S) \supset H(F, do(A, S)) \equiv \\ & \qquad [\mathit{Caused}(F, \mathit{true}, do(A, S)) \vee H(F, S) \wedge \neg \mathit{Caused}(F, \mathit{false}, do(A, S))]. \end{aligned}$$

Since $M$ is a model of $\mathit{Poss}(A, S)$, we have

$$\begin{aligned} & H(F, do(A, S)) \equiv \\ & \qquad [\mathit{Caused}(F, \mathit{true}, do(A, S)) \vee H(F, S) \wedge \neg \mathit{Caused}(F, \mathit{false}, do(A, S))]. \end{aligned} \tag{8}$$

Now consider the circumscription of $\mathit{Caused}$ in $\mathcal{T}$ with all other predicates fixed. Notice that all axioms about $\mathit{Caused}$ in $\mathcal{T}$ have the form $W \supset \mathit{Caused}(x, y, z)$, where $W$ is a formula that does not mention $\mathit{Caused}$. Therefore the circumscription of $\mathit{Caused}$ is equivalent to the predicate completion of $\mathit{Caused}$. Suppose $F$ is $f(t)$, and that all of the axioms about $\mathit{Caused}(f(x), v, s)$ in $\mathcal{T}$ are as follows:

$$W_1 \supset \mathit{Caused}(f(x), v, s), \cdots, W_i \supset \mathit{Caused}(f(x), v, s).$$

Because of the unique names axioms about fluents, the result of predicate completion on $\mathit{Caused}$ will entail:

$$\mathit{Caused}(f(x), v, s) \equiv W_1 \vee \cdots \vee W_i.$$

Now $W_1$,...,$W_i$ are from action effect axioms and domain rules about $f$. By the way that (1) is generated, and noting that $\mathit{Action}(A)$, $\mathit{Fluent}(F)$, and $\mathit{Poss}(A, S)$ are true, one can see that when the above equivalence is instantiated by replacing $t$ for $x$, $\mathit{true}$ for $v$, and $S$ for $s$, we will get

$$\begin{aligned} \mathit{Caused}(F, \mathit{true}, do(A, S)) \equiv\ & H(\varphi_1, S) \vee \cdots \vee H(\varphi_n, S) \vee \\ & H(\varphi'_1, do(A, S)) \vee \cdots \vee H(\varphi'_l, do(A, S)). \end{aligned}$$

Similarly, we have the following axiom about $Caused(F, false, do(A, S))$:

$$Caused(F, false, do(A, S)) \equiv H(\phi_1, S) \vee \cdots \vee H(\phi_m, S) \vee \\ H(\phi'_1, do(A, S)) \vee \cdots \vee H(\phi'_k, do(A, S)).$$

From these two axioms and (8), we get:

$$\begin{aligned} H(F, do(A, S)) \quad \equiv \quad & H(\varphi_1, S) \vee \cdots \vee H(\varphi_n, S) \vee \\ & H(\varphi'_1, do(A, S)) \vee \cdots \vee H(\varphi'_l, do(A, S)) \vee \\ & H(F, S) \wedge \neg[H(\phi_1, S) \vee \cdots \vee H(\phi_m, S) \vee \\ & \qquad H(\phi'_1, do(A, S)) \vee \cdots \vee H(\phi'_k, do(A, S))]. \end{aligned}$$

Since $M$ is a model of $Comp(\mathcal{T})$, $M$ satisfies the above formula. By the construction of $M_{S,A}$, it satisfies the pseudo-successor state axiom (1).

Finally, the fact that $M_{S,A}$ is a model of $Succ1$ should be apparent. □

In general, (6) does not imply (7). There are several reasons:

- As we mentioned after the procedure CCP, we assume that all information about the initial situation is given by $Init$.
- Our procedure works on actions one at a time. The situation calculus theory $\mathcal{T}$ captures the effects of all actions in a single theory. So it is possible that a bad specification of an action causes the entire theory to become inconsistent. For instance, if we have $Causes(true, p)$, $Precond(A, true)$, and $Effect(A, false, p)$, then the corresponding situation calculus theory will be inconsistent because of action $A$. But for our procedure, it will generate an inconsistent theory only on $A$.

## 5. An Implementation

Except for Step 2.2, the procedure CCP in Section 3 is straightforward to implement. This section describes the strategy that our system uses for implementing Step 2.2. The main idea comes from the work of Lin (2001a) on strongest necessary and weakest sufficient conditions.

Given a propositional theory $T$, a proposition $q$, and a set $B$ of propositions, a formula $\varphi$ is said to be a sufficient condition of $q$ on $B$ under $T$ if $\varphi$ consists of propositions in $B$ and $T \models \varphi \supset q$. It is said to be the *weakest* sufficient condition if for any such sufficient condition $\varphi'$, we have that $T \models \varphi' \supset \varphi$. Similarly, a formula $\varphi$ is said to be a necessary condition of $q$ on $B$ under $T$ if $\varphi$ consists of propositions in $B$ and $T \models q \supset \varphi$. It is said to be the *strongest* necessary condition if for any such necessary condition $\varphi'$, we have that $T \models \varphi \supset \varphi'$.

It is easy to see that the weakest sufficient condition and the strongest necessary condition are unique up to logical equivalence under the background theory. It was shown (Lin, 2001a) that these two notions are closely related, and can be computed using the technique of forgetting (Lin & Reiter, 1994a). In particular, for action theories, an effective strategy is to first compute the strongest necessary condition, add it to the background theory, and then compute the weakest sufficient condition under the new theory. This strategy is justified by the following proposition Lin (2001a):

**Proposition 1** *Let $T$ be a theory, $q$ a proposition, and $B$ a set of propositions. If $\varphi$ is a necessary condition of $q$ on $B$ under $T$, and $\psi$ the weakest sufficient condition of $q$ on $B$ under $T \cup \{\varphi\}$, then $\varphi \wedge \psi$ is the weakest sufficient condition of $q$ on $B$ under $T$.*

We can now describe our strategy for implementing Step 2.2 of the procedure CCP. In the following, given an action instance $A$, as in Step 2.2, let $Succ$ be the set of pseudo-successor state axioms for primitive fluent atoms, $Succ1$ the set of pseudo-successor state axioms for complex fluent atoms, and $Init$ the set of initial situation axioms derived from the action precondition axiom for $A$, domain axioms, domain rules, and complex fluent definitions. Also in the following, a *succ*-proposition is one of the form $succ(f)$, and an *init*-proposition is one of the form $init(f)$.

1. Transform $Init$ into a clausal form and derive from it a set of unit clauses $Unit$.

2. Use $Unit$ to simplify the axioms in $Succ$ and for each resulting axiom in it:

$$succ(f) \equiv \Phi_f, \tag{9}$$

   if $\Phi_f$ does not mention *succ*-propositions, then delete it from $Succ$, output it and replace $succ(f)$ in the rest of the axioms by $\Phi_f$.

3. For each fluent atom $f$ whose pseudo-successor state axiom (9) is in $Succ$, if $\Phi_f$ has the form $init(f) \wedge ...$ (a candidate of a frame axiom), then check to see if $succ(f)$ can be derived from $Succ$, $Unit$, and $init(f)$ by unit resolution. If so, delete it from $Succ$, output $succ(f) \equiv init(f)$, and replace $succ(f)$ in $Succ$ by $init(f)$.

4. For each fluent atom $f$ whose pseudo-successor state axiom (9) is in $Succ$, compute the strongest necessary condition $\alpha_f$ of $succ(f)$ on the *init*-propositions under the theory $Init \cup Succ$, and the weakest sufficient condition $\beta_f$ of $succ(f)$ on the *init*-propositions under the theory $\{\alpha_f\} \cup Init \cup Succ$. If $\beta_f$ is a tautology, then delete (9) from $Succ$, output $succ(f) \equiv \alpha_f$, and replace $succ(f)$ in $Succ$ by $\alpha_f$. If $\beta$ is not a tautology, then output $succ(f) \supset \alpha_f$ and $\alpha_f \wedge \beta_f \supset succ(f)$, but do not delete (9) from $Succ$. The correctness of this step follows from Proposition 1.

5. The previous steps solve the equations in $Succ$, and generate appropriate output for primitive fluent atoms. For each complex fluent atom $F$:

$$Defined(F, \varphi),$$

   if every primitive fluent atom in $\varphi$ has a successor state axiom, then do the following:

   (a) if no primitive fluent atoms in $\varphi$ are changed by the action, then this complex fluent atom is not changed by the action either, so output $succ(F) \equiv init(F)$;

   (b) otherwise, output $succ(F) \equiv \Phi$, where $\Phi$ is obtained from $succ(\varphi)$ by replacing every *succ*-proposition in it by the right side of its successor state axiom.

Otherwise, if some of the primitive fluent atoms in $\varphi$ do not have a successor state axiom, which means the action may have an indeterminate effect on them, then this action may have an indeterminate effect on $F$ as well. Compute the strongest necessary and weakest sufficient conditions of $succ(F)$ under $Init \cup Succ \cup Succ1$ as in the last step, and output them.

6. This step will try to generate a STRIPS-like description for the action instance $A$ based on the results of Steps 4 and 5. For each fluent atom $F$, do according to one of the following cases:

   (a) if its successor state axiom is $succ(F) \equiv true$, then put $F$ into the add list unless $init(F)$ is entailed by $Init$;

   (b) if its successor state axiom is $succ(F) \equiv false$, then put $F$ into the delete list unless $\neg init(F)$ is entailed by $Init$;

   (c) if its successor state axiom is $succ(F) \equiv \Phi$, and $\Phi$ is not the same as $true$, $false$, or $init(F)$, then put $F$ in the conditional effect list and output its successor state axiom.

   (d) If $F$ does not have a successor state axiom, then put it in the list of indeterminate effects.

   Clearly, if $F$ is not put into any of the lists, then its truth value is not affected by $A$.

Steps 4 and 5 of the above procedure are the bottleneck as in the worst case, computing the strongest necessary condition of a proposition is coNP-hard. However, it has been our experience that if action $A$ has a context-free effect on fluent atom $F$, then its successor state axiom can be computed without going through Step 4.

We have implemented the procedure CCP using the above strategy in SWI-Prolog 3.2.9[2]. The url for this system is as follows:

`http://www.cs.ust.hk/~flin/ccp.html`

Using the system, we have encoded in our action description language many of the planning domains that come with the original release of PDDL (McDermott, 1998), and compiled them to STRIPS-like specifications. Our encodings of the domains and the results returned by the system are included in the online appendix. In the following, we illustrate some interesting features of our system using the following two domains: the blocks world and the monkey and bananas domain.

### 5.1 The Blocks World

We have used the blocks world as the running example. Here we shall give an alternative specification of the domain using the following better known set of actions: *stack*, *unstack*, *pickup*, and *putdown*. We shall use this domain to show that changing slightly the precondition of one of the actions can result in a very different action specification.

2. SWI-Prolog is developed by Jan Wielemaker at University of Amsterdam

We begin with a description that corresponds to the standard STRIPS encoding of the domain.

$$
\begin{aligned}
&\mathit{Fluent}(on(x,y), block(x) \wedge block(y)),\\
&\mathit{Fluent}(ontable(x), block(x)),\\
&\mathit{Fluent}(holding(x), block(x)),\\
&\mathit{Complex}(clear(x), block(x),\\
&\mathit{Defined}(clear(x), (\neg\exists(y, block) on(y,x)) \wedge \neg holding(x)),\\
&\mathit{Complex}(handempty, true),\\
&\mathit{Defined}(handempty, \neg\exists(x, block) holding(x)),
\end{aligned}
$$

$$
\begin{aligned}
&\mathit{Causes}(on(x,y) \wedge x \neq z, \neg on(z,y)),\\
&\mathit{Causes}(on(x,y) \wedge y \neq z, \neg on(x,z)),\\
&\mathit{Causes}(on(x,y), \neg ontable(x)),\\
&\mathit{Causes}(ontable(x), \neg on(x,y)),\\
&\mathit{Causes}(on(x,y), \neg holding(x)),\\
&\mathit{Causes}(on(x,y), \neg holding(y)),\\
&\mathit{Causes}(holding(x), \neg ontable(x)),\\
&\mathit{Causes}(holding(x), \neg on(x,y)),\\
&\mathit{Causes}(holding(x), \neg on(y,x)),\\
&\mathit{Causes}(holding(x) \wedge y \neq x, \neg holding(y)),
\end{aligned}
$$

$$
\begin{aligned}
&\mathit{Action}(stack(x,y), block(x) \wedge block(y) \wedge x \neq y),\\
&\mathit{Precond}(stack(x,y), holding(x) \wedge clear(y)),\\
&\mathit{Effect}(stack(x,y), true, on(x,y)),
\end{aligned}
$$

$$
\begin{aligned}
&\mathit{Action}(unstack(x,y), block(x) \wedge block(y) \wedge x \neq y),\\
&\mathit{Precond}(unstack(x,y), clear(x) \wedge on(x,y) \wedge handempty),\\
&\mathit{Effect}(unstack(x,y), true, holding(x)),
\end{aligned}
$$

$$
\begin{aligned}
&\mathit{Action}(putdown(x), block(x)),\\
&\mathit{Precond}(putdown(x), holding(x)),\\
&\mathit{Effect}(putdown(x), true, ontable(x)),
\end{aligned}
$$

$$
\begin{aligned}
&\mathit{Action}(pickup(x), block(x)),\\
&\mathit{Precond}(pickup(x), handempty \wedge ontable(x) \wedge clear(x)),\\
&\mathit{Effect}(pickup(x), true, holding(x)).
\end{aligned}
$$

Notice that compared to the description in Example 1, there are two more fluents, *holding* and *handempty* here. Thus we have a few more domain rules about them, and the definition of *clear* is changed to take into account that when a block is held, it is not considered to be clear.

Now assuming a domain with three blocks $Domain(block, \{1, 2, 3\})$, our system will generate 19 fluent atoms, and 18 action instances. For each action instance, it returns both a complete set of successor state axioms and a STRIPS-like representation. The total computation time for all actions is 835K inferences and 0.5 seconds.[3] This is a pure STRIPS domain, i.e. all actions are context free. For this type of domains, as we mentioned earlier, Step 4 in our implementation procedure is not needed, and Step 5 is easy.

The results are as expected. For instance, for action *pickup*(1), the STRIPS-like representation returned by the system looks like the following: track 1 to track 2), the STRIPS-like representation looks like:

```
pickup(1):
  Preconditions: clear(1), handempty, ontable(1)
  Add list: holding(1)
  Delete list: ontable(1), clear(1), handempty
  Conditional effects:
  Indeterminate effects:
```

The complete output is given in the online appendix. Now let us consider what happen if we drop *ontable*($x$) from the precondition of *pickup*($x$):

$$Precond(pickup(x), handempty \wedge clear(x)).$$

This means that as long as a block is clear, it can be picked up. With this new precondition, our system returns the following STRIPS-like representation for action *pickup*(1);

```
pickup(1):
  Preconditions: clear(1), handempty
  Add list: holding(1)
  Delete list: clear(1), handempty, on(1, 2), on(1, 3), ontable(1)
  Conditional effects:
     succ(clear(2))<-> - (init(on(2, 2))\/init(on(3, 2)))
     succ(clear(3))<-> - (init(on(2, 3))\/init(on(3, 3)))
  Indeterminate effects:
```

Here "-" is negation, and "\/" is disjunction. An ADL-like description for this action would be something like the following:

```
pickup(x):
  Preconditions: clear(x), handempty
  Add list: holding(x),
            clear(y) when on(x,y)
  Delete list: clear(x), handempty,
               on(x,y) when on(x,y)
               ontable(x) when ontable(x)
```

3. All times in this paper refer to CPU times on a Pentium III 1GHz machine with 512MB RAM running SWI-Prolog 3.2.9 under Linux. The number of inferences is the one reported by SWI-Prolog, and roughly corresponds to the number of resolution steps carried out by the Prolog interpreter, and is machine independent.

### 5.2 The Monkey and Bananas Domain

This domain is again adapted from McDermott's PDDL library of planning domains, which attributes it to the University of Washington's UCPOP collection of action domains, which in turn attributes it to Prodigy. While some of the action effects generated by our system are context-dependent, they are all context-free in the other systems. We shall elaborate on this difference later.

In this domain, there are two types, *loc* for locations (we assume there are three locations here), and *object* for things like monkey, banana, box, etc.:

$$Domain(loc, \{1, 2, 3\}),$$
$$Domain(object, \{monkey, box, banana, knife, glass, fountain\}).$$

The following are fluent definitions:

$$Fluent(onFloor),$$
$$Fluent(at(M, X), object(M) \wedge loc(X)),$$
$$Fluent(hasknife),$$
$$Fluent(onbox(X), loc(X)),$$
$$Fluent(hasbanana),$$
$$Fluent(haswater),$$
$$Fluent(hasglass).$$

The following are domain rules about these fluents:

$$Causes(onbox(X), at(monkey, X)), \quad (10)$$
$$Causes(onbox(X), at(box, X)), \quad (11)$$
$$Causes(onbox(X), \neg onFloor), \quad (12)$$
$$Causes(onFloor, \neg onbox(X)), \quad (13)$$
$$Causes(at(M, X) \wedge X \neq Y, \neg at(M, Y)), \quad (14)$$
$$Causes(hasglass \wedge at(monkey, X), at(glass, X)), \quad (15)$$
$$Causes(hasknife \wedge at(monkey, X), at(knife, X)), \quad (16)$$
$$Causes(hasbanana \wedge at(monkey, X), at(banana, X)). \quad (17)$$

The following are action definitions along with their respective preconditions and effect axioms:

- $goto(x, y)$ - the monkey goes to $x$ from $y$:

$$Action(goto(X, Y), loc(X) \wedge loc(Y) \wedge X \neq Y),$$
$$Precond(goto(X, Y), at(monkey, Y) \wedge onFloor),$$
$$Effect(goto(X, Y), true, at(monkey, X)).$$

- $climb(X)$ - the monkey climbs onto the box at location $X$:

$$Action(climb(X), loc(X)),$$
$$Precond(climb(X), at(box, X) \wedge onFloor \wedge at(monkey, X)),$$
$$Effect(climb(X), true, onbox(X)).$$

- $pushbox(X, Y)$ - the monkey pushes the box from $Y$ to $X$.

$$Action(pushbox(X, Y), loc(X) \wedge loc(Y) \wedge X \neq Y),$$
$$Precond(pushbox(X, Y), at(monkey, Y) \wedge at(box, Y)) \wedge onFloor),$$
$$Effect(pushbox(X, Y), true, at(monkey, X)),$$
$$Effect(pushbox(X, Y), true, at(box, X)).$$

- $getknife(X)$ - get knife at location $X$.

$$Action(getknife(X), loc(X)),$$
$$Precond(getknife(X), at(knife, X) \wedge at(monkey, X) \wedge \neg hasknife),$$
$$Effect(getknife(X), true, hasknife).$$

- $getbanana(X)$ - grab banana at loc $X$, provided the monkey is on the box.

$$Action(getbanana(X), loc(X)),$$
$$Precond(getbanana(X), onbox(X) \wedge at(banana, X) \wedge \neg hasbanana),$$
$$Effect(getbanana(X), true, hasbanana).$$

- $pickglass(X)$ - pick up glass at loc $X$.

$$Action(pickglass(X), loc(X)),$$
$$Precond(pickglass(X), at(glass, X) \wedge at(monkey, X) \wedge \neg hasglass),$$
$$Effect(pickglass(X), true, hasglass).$$

- $getwater(X)$ - get water from fountain at loc $X$, provided the monkey is on the box, and has a glass in hand.

$$Action(getwater(X), loc(X)),$$
$$Precond(getwater(X), at(fountain, X) \wedge onbox(X) \wedge hasglass \wedge \neg haswater),$$
$$Effect(getwater(X), true, haswater).$$

This domain has 27 actions and 26 fluent atoms. Again, for each action, our system generates both a complete set of fully instantiated successor state axioms and a STRIPS-like representation. For instance, for action $goto(1, 2)$, the following is the STRIPS-like representation generated by the system:

```
Action goto(1, 2)

Preconditions: at(monkey, 2), onFloor

Add list: at(monkey, 1)

Delete list: at(monkey, 2)

Conditional effects:

succ(at(banana, 1)) <-> init(hasbanana) \/ init(at(banana, 1))
succ(at(knife, 1)) <-> init(hasknife) \/ init(at(knife, 1))
succ(at(glass, 1)) <-> init(hasglass) \/ init(at(glass, 1))
succ(at(banana, 2)) <-> - init(hasbanana) & init(at(banana, 2))
succ(at(knife, 2)) <-> - init(hasknife) & init(at(knife, 2))
succ(at(glass, 2)) <-> - init(hasglass) & init(at(glass, 2))
```

The total running time for all actions is 8 seconds while performing 20 million inferences. About 90 percent of time is spent on Step 4, i.e. on computing the strongest necessary and weakest sufficient conditions of fluent atoms on which the given action has context-dependent effects. For instance, for action $goto(1, 2)$ above, the majority of time was spent on generating the above 6 conditional effects.

For this action, actually for all the actions in this domain, we could use an ADL-like description (Pednault, 1989) for conditional effects:

```
Add list: at(banana,1) when hasbanana
          at(knife,1) when hasknife
          at(glass,1) when hasglass

Delete list: at(banana,2) when hasbanana
             at(knife,2) when hasknife
             at(glass,2) when hasglass
```

However, it is not clear whether this can always be done in the general case.

We mentioned earlier that the specifications for this domain given in McDermott's collection as well as others are all context-free. For instance, the following is a specification for action *goto* in PDDL in McDermott's collection:

```
(:action GO-TO
          :parameters (?x ?y)
          :precondition (and (location ?x) (location ?y)
                  (not (= ?y ?x)) (on-floor) (at monkey ?y))
          :effect (and (at monkey ?x) (not (at monkey ?y))))
```

This corresponds to a context-free action that does not change any other fluent except *at*. It is clear that the design of this action does not take into account domain rules (15) - (17). With this specification, if initially banana is at location 1, then the goal of having banana at location 2 would not be achievable.

### 5.3 Summary

The other domains that we have experimented including a scheduling domain that includes Pednault's dictionary and paycheck domain as a special case, the rocket domain, the SRI robot domain, the machine shop assembling domain, the ferry domain, the grid domain, the sokoban domain, and the gear domain. They are all included in the online appendix. We summarize below some of the common features of these domains:

- In all the domains that we tried, it is quite straightforward to decide what effects of an action should be encoded as direct effects (those given by the predicate *Effect*) and what effects as indirect effects (those derived from domain rules).
- The most common domain rules are functional dependency constraints. For instance, in the blocks world, the fluent atom $on(x, y)$ is functional on both arguments; in the monkey and banana domain, the fluent atom $at(object, loc)$ is functional on the second argument (each object can be at only one location). It makes sense then that we would have a special shorthand for these domain rules, and perhaps a special procedure for handling them as well. But more significantly, given the prevalence of these functional dependency constraints in action domains, it is worthwhile to investigate the possibility of a general purpose planner making good use of these constraints.
- As we mentioned earlier, our system is propositional. The generated successor state axioms and STRIPS-like systems are all fully instantiated. However, it is often easy for the user to generalize these propositional specifications to first-order ones. We shall investigate the generality of this observation next.

## 6. Generalizing Propositional STRIPS-Like Systems to Ones With Parameters

As we mentioned, for many action domain descriptions, the successor state axioms and STRIPS-like systems generated for a specific domain can be generalized to arbitrary ones.

More precisely, let $\mathcal{D}$ be a domain description, and

$$Domain(p_1, D_{p_1}), \cdots, Domain(p_k, D_{p_k})$$

its type specification. Suppose in $\mathcal{D}$ for action $A$ we have that $Init_A \cup Succ_A \models \varphi$. Now suppose $\mathcal{D}'$ is another domain description that is just like $\mathcal{D}$ except that it has a different type specification:

$$Domain(p_1, D'_{p_1}), \cdots, Domain(p_k, D'_{p_k}).$$

The question that we are interested in is this: given any one-to-one mapping from the type specification of $\mathcal{D}$ to that of $\mathcal{D}'$, will $Init_{A'} \cup Succ_{A'} \models \varphi'$ be true in $\mathcal{D}'$? Here $A'$ (resp. $\varphi'$) is the result of replacing all objects in $A$ (resp. $\varphi$) according to the mapping.

For instance, if the above is true for the blocks world, then we can generalize the results for the domain description in Example 1 as follows. As we have shown, for action $stack(1, 2)$, both $succ(on(1, 2))$ and $\neg succ(on(1, 3))$ are true. Now if we change the type specification to $Domain(block, \{a, b, c, d, e\})$, and if we map 1 to $a$, 2 to $c$, and 3 to $e$, in the new domain specification, we will have that for action $stack(a, c)$, $succ(on(a, c))$ and $\neg succ(on(a, e))$ are true. Furthermore, by changing the mapping for 3, we see that for any $x$ that is different from $a$ and $c$ (the mapping needs to be one-to-one), $\neg succ(on(a, x))$ is true.

Obviously, this is to be expected of the blocks world. We now proceed to show that for some general classes of domain descriptions, we can do this as well. We first make precise the mapping from one type specification to another.

**Definition 2** *Given two type specifications $O$:*

$$Domain(p_1, D_{p_1}), \cdots, Domain(p_k, D_{p_k}),$$

*and $O'$:*

$$Domain(p_1, D'_{p_1}), \cdots, Domain(p_k, D'_{p_k}),$$

*an* embedding *from $O$ to $O'$ is a one-to-one mapping $\tau$ from $D_{p_1} \cup \cdots \cup D_{p_k}$ to $D'_{p_1} \cup \cdots \cup D'_{p_k}$ such that for any $1 \leq i \leq k$, and any $a \in D_{p_i}$, $f(a) \in D'_{p_i}$.*

Clearly, if there is an embedding of $O$ to $O'$, then for each type $p$, the size of the domain for $p$ in $O'$ must be at least the size of the domain for $p$ in $O$. Given such an embedding $\tau$, any expression $\beta$ (actions, propositions, formulas) in an action domain description $\mathcal{D}$ with $O$ as its type specification can be mapped to $\tau(\beta)$ in the language of $\mathcal{D}'$: one simply replaces each object $a$ in $\beta$ by $\tau(a)$, where $\mathcal{D}'$ differs from $\mathcal{D}$ only in that it uses $O'$ as its type specification. Notice that only objects (those in the domain of some type) are to be replaced, not constants that may occur in the effect axioms or domain rules.

**Definition 3** *An action domain description belongs to* simple-I *class if it does not mention any function of positive arity, does not mention any complex fluents except in complex fluent definitions, and satisfies the following conditions:*

1. *If $Precond(A, \phi_A)$ is an action precondition definition, then $\phi_A$ has the form $(\forall x, p)...(\forall y, q)W$, where $W$ is a fluent formula that does not have any quantifiers.*

2. *If $Effect(A, \varphi, F)$ or $Effect(A, \varphi, \neg F)$ is an action effect axiom, then $\varphi$ does not have any quantifiers, and the variables in $\varphi$ and $F$ are among those in $A$. That is, one cannot have something like*

$$Effect(explodeAt(x), nearby(y, x), dead(y)).$$

3. *If $Causes(\varphi, F)$ or $Causes(\varphi, \neg F)$ is a domain rule, then $\varphi$ does not have any quantifiers, and all the variables in $\varphi$ must be in $F$.*

**Theorem 2** *Let $\mathcal{D}$ be a simple-I action domain description, and $A$ an action instance in $\mathcal{D}$. Let $\mathcal{D}'$ be just like $\mathcal{D}$ except for the type specification. Then for any formula $\psi$ that does not mention any complex fluent and has no quantifiers, and any embedding $\tau$ from the type specification of $\mathcal{D}$ to that of $\mathcal{D}'$, we have that if $Init_A \cup Succ_A \models \psi$ in $\mathcal{D}$. then $Init_{\tau(A)} \cup Succ_{\tau(A)} \models \tau(\psi)$ in $\mathcal{D}'$.*

**Proof:** Suppose $Init_{\tau(A)} \cup Succ_{\tau(A)} \models \tau(\psi)$ is not true, and that $M_1$ is a truth assignment in the language of $\mathcal{D}'$ that satisfies $Init_{\tau(A)} \cup Succ_{\tau(A)}$ and $\neg\tau(\psi)$. Now construct a truth assignment $M_2$ in the language of $\mathcal{D}$ as follows: for any proposition $P$ in the language of $\mathcal{D}$ that does not mention any complex fluent, $M_2 \models P$ iff $M_1 \models \tau(P)$ ($P$ is really either a static proposition, $succ(F)$,or $init(F)$, where $F$ is a primitive fluent atom). The truth values of complex fluent atoms in $M_2$ are defined according to their definitions. Clearly, $M_2 \models \psi$. We now need to show that $M_2$ also satisfies $Init_A$ and $Succ_A$. For $Init_A$, there are three cases:

1. $M_2 \models init(F) \equiv init(\varphi)$ when $Defined(F, \varphi)$ is a complex fluent definition. This follows from the construction of $M_2$.

2. $M_2 \models init(\phi_A)$ when $Precond(A, \phi_A)$ is the precondition definition for $A$. By our assumption, $\phi_A$ has the form $(\forall x, p)...(\forall y, q).W$, where $W$ is a formula without any quantifiers. Without loss of generality, let us assume it is $(\forall x, p)W$. Then this formula is equivalent to
$$\bigvee_{a \in D_p} W(x/a)$$
under $\mathcal{D}$, where $D_p$ is the domain of type $p$ in $\mathcal{D}$. So $M_2 \models (\forall x, p)W$ iff
$$M_2 \models \bigvee_{a \in D_p} W(x/a)$$
iff
$$M_1 \models \bigvee_{a \in D_p} W(x/\tau(a)),$$
which is true since $M_1 \models (\forall x, p)W$.

3. All other formulas in $Init_A$ do not mention complex fluents and have no quantifiers. They are true in $M_2$ because the corresponding ones are true in $M_1$.

For $Succ_A$, suppose $F$ is a primitive fluent atom, and its pseudo-successor state axiom $\Phi_F$ as constructed according to the procedure CCP given in Section 3 is as follows:

$$succ(F) \equiv init(\varphi_1) \vee \cdots \vee init(\varphi_n) \vee succ(\varphi_1') \vee \cdots \vee succ(\varphi_l') \vee \\ init(F) \wedge \neg[init(\phi_1) \vee \cdots \vee init(\phi_m) \vee succ(\phi_1') \vee \cdots \vee succ(\phi_k')].$$

Because of the following properties about $\mathcal{D}$:

- each effect axiom $Effect(A, \varphi, F)$ or $Effect(A, \varphi, \neg F)$ has the property that $\varphi$ has no quantifier, and that the variables in $\varphi$ are also in $F$;

- each domain rule of the form $Causes(\varphi, F)$ or $Causes(\varphi, \neg F)$ has the property that $\varphi$ has no quantifier, and that the variables in $\varphi$ are also in $F$;

so the pseudo-successor state axiom for $\tau(succ(F))$ under $\mathcal{D}'$ is just $\tau(\Phi_F)$. Thus $M_2 \models \Phi_F$ since $M_1 \models \tau(\Phi_F)$. This proves that $M_2$ is a model of $Succ_A$, thus the theorem. □

However, most of the examples that we have in the paper do not belong to this simple-I class, for two reasons: action preconditions, like those in the blocks world, can mention complex fluents; and some of the negative domain rules $Causes(\varphi, \neg F)$ may have some variables not in $F$. The first problem is not a problem in principle as complex fluents can be replaced by their definitions. The second problem is more serious, and that leads to a new type of simple action theories.

**Definition 4** *An action domain description belongs to* simple-II *class if it does not mention any function of positive arity, does not mention any complex fluents except in complex fluent definitions, and satisfies the following conditions:*

1. *If $Precond(A, \phi_A)$ is an action precondition definition, then $\phi_A$ has the form $(\forall x, p)...(\forall y, q)W$, where $W$ is a fluent formula that does not have any quantifiers.*
2. *If $Effect(A, \varphi, F)$ or $Effect(A, \varphi, \neg F)$ is an action effect axiom, then $\varphi$ does not have any quantifiers, and the variables in $\varphi$ and $F$ are among those in $A$.*
3. *There are no positive domain rules of the form $Causes(\varphi, F)$.*
4. *If $Causes(\varphi, \neg F)$ is a domain rule, then $\varphi$ must be of the form $\varphi_1 \wedge \varphi_2$, where $\varphi_1$ is any formula that does not mention any fluents and $\varphi_2$ is a fluent atom. Notice that there is no restriction on variables in $\varphi_2$.*

Simple-II class action domain descriptions seem to be very limited in that there can be no positive domain rules, and the only negative domain rules allowed are binary. Nevertheless, they still capture many context-free action domains. For instance, both the blocks world and meet-and-pass domains in this paper belong to this class: for the blocks world, notice that while it uses the complex fluent *clear* in some of its action precondition definitions, as in $Precond(stack(x, y), ontable(x) \wedge clear(x) \wedge clear(y))$, these definitions can be reformulated as follows using *clear*'s definition:

$$Precond(stack(x, y),\\ ontable(x) \wedge (\forall x_1, block)(\forall y_1, block)(\neg on(x_1, x) \wedge \neg on(y_1, y))).$$

This will then satisfy the condition about *Precond* in the above definition of simple-II action domain descriptions. While we have not verified it formally, it seems that all the context-free action domains in McDermott's PDDL library of action domains, including the logistics domain, belong to the simple-II class.

**Theorem 3** *Let $\mathcal{D}$ be a simple-II action domain description, and $A$ an action instance in $\mathcal{D}$. Let $\mathcal{D}'$ be just like $\mathcal{D}$ except for the type specification. Then for any formula $\psi$ that does not mention any complex fluent and has no quantifiers, and any embedding $\tau$ from the type specification of $\mathcal{D}$ to that of $\mathcal{D}'$, we have that if $Init_A \cup Succ_A \models \psi$ in $\mathcal{D}$ then $Init_{\tau(A)} \cup Succ_{\tau(A)} \models \tau(\psi)$ in $\mathcal{D}'$.*

**Proof:** Suppose $Init_{\tau(A)} \cup Succ_{\tau(A)} \models \tau(\psi)$ is not true, and that $M_1$ is a truth assignment in the language of $\mathcal{D}'$ that satisfies $Init_{\tau(A)} \cup Succ_{\tau(A)}$ and $\neg\tau(\psi)$. Now construct a truth assignment $M_2$ in the language of $\mathcal{D}$ as follows: for any proposition $P$ in the language of $\mathcal{D}$ that does not mention any complex fluent, $M_2 \models P$ iff $M_1 \models \tau(P)$ ($P$ is really either a static proposition, $succ(F)$,or $init(F)$, where $F$ is a primitive fluent atom). The truth values of complex fluent atoms in $M_2$ are defined according to their definitions. Clearly, $M_2 \models \psi$. We now need to show that $M_2$ also satisfies $Init_A$ and $Succ_A$. For $Init_A$, there are three cases:

1. $M_2 \models init(F) \equiv init(\varphi)$ when $Defined(F, \varphi)$ is a complex fluent definition. This follows from the construction of $M_2$.

2. $M_2 \models init(\phi_A)$ when $Precond(A, \phi_A)$ is the precondition definition for $A$. By our assumption, $\phi_A$ has the form $(\forall x, p)...(\forall y, q).W$, where $W$ is a formula without any quantifiers. Without loss of generality, let us assume it is $(\forall x, p_1)W$. Then this formula is equivalent to
$$W(x/a_{11}) \vee \cdots \vee W(x/a_{1n_1})$$
under $\mathcal{D}$. So $M_2 \models (\forall x, p_1)W$ iff
$$M_2 \models W(x/a_{11}) \vee \cdots \vee W(x/a_{1n_1})$$
iff
$$M_1 \models W(x/\tau(a_{11})) \vee \cdots \vee W(x/\tau(a_{1n_1})),$$
which is true since $M_1 \models (\forall x, p_1)W$.

3. All other formulas in $Init_A$ do not mention complex fluents and have no quantifiers. They are true in $M_2$ because the corresponding ones are true in $M_1$.

For $Succ_A$, suppose $F$ is a primitive fluent atom. Since there is no positive domain rule of the form $Causes(\varphi, F)$, the pseudo-successor state axiom for $F$ as constructed according to the procedure CCP given in Section 3 must be of the following form:

$$\begin{aligned} &succ(F) \equiv init(\varphi_1) \vee \cdots \vee init(\varphi_n) \vee \\ &\quad init(F) \wedge \neg[init(\phi_1) \vee \cdots \vee init(\phi_m) \vee succ(\phi'_1) \vee \cdots \vee succ(\phi'_k)], \end{aligned}$$

where for each $1 \leq i \leq k$, $Causes(\phi'_i, \neg F)$ is an instance of a domain rule in $\mathcal{D}$.

Because in $\mathcal{D}$ and $\mathcal{D}'$, each effect axiom $Effect(\varphi, F)$ or $Effect(\varphi, \neg F)$ has the property that $\varphi$ has no quantifier, and that the variables in $\varphi$ are also in $F$, the pseudo-successor state axiom for $\tau(succ(F))$ under $\mathcal{D}'$ must have the form:

$$\begin{aligned} &succ(\tau(F)) \equiv init(\tau(\varphi_1)) \vee \cdots \vee init(\tau(\varphi_n)) \vee \\ &\quad init(\tau(F)) \wedge \neg[init(\tau(\phi_1)) \vee \cdots \vee init(\tau(\phi_m)) \vee \\ &\quad succ(\tau(\phi'_1)) \vee \cdots \vee succ(\tau(\phi'_k)) \vee succ(\phi)], \end{aligned} \tag{18}$$

where $\phi$ is a disjunction such that each disjunct $\alpha$ must be such that $Causes(\alpha, \neg\tau(F))$ is an instance in $\mathcal{D}'$ and that the fluent atom in $\alpha$ contains an object not in $\tau(A)$ and $\tau(F)$.

There are two cases:

- Suppose $M_2 \models succ(F)$. Then $M_1 \models succ(\tau(F))$. Since $M_1$ is a model of $Succ_{\tau(A)}$, $M_1$ satisfies the above axiom about $succ(\tau(F))$. Therefore $M_1$ satisfies the following formula:

$$\begin{aligned}&init(\tau(\varphi_1)) \vee \cdots \vee init(\tau(\varphi_n)) \vee \\ &\quad init(\tau(F)) \wedge \neg[init(\tau(\phi_1)) \vee \cdots \vee init(\tau(\phi_m)) \vee \\ &\qquad succ(\tau(\phi'_1)) \vee \cdots \vee succ(\tau(\phi'_k))].\end{aligned}$$

  Since the above formula does not mention any complex fluents and has no quantifiers, $M_2$ satisfies the corresponding formula:

$$\begin{aligned}&init(\varphi_1) \vee \cdots \vee init(\varphi_n) \vee \\ &\quad init(F) \wedge \neg[init(\phi_1) \vee \cdots \vee init(\phi_m) \vee succ(\phi'_1) \vee \cdots \vee succ(\phi'_k)],\end{aligned} \tag{19}$$

  which is the right side of the equivalence of the pseudo-successor state axiom for $succ(F)$ in $Succ_A$.

- Now suppose $M_2$ satisfies (19). We'll show that $M_1$ satisfies the right side of (18), thus $M_1 \models succ(\tau(F))$ so $M_2 \models succ(F)$. There are two cases:

  - $M_2$ satisfies the following formula:

$$init(\varphi_1) \vee \cdots \vee init(\varphi_n). \tag{20}$$

    In this case, since the above formula does not mention any complex fluents and has no quantifier, $M_1$ satisfies the following corresponding formula:

$$init(\tau(\varphi_1)) \vee \cdots \vee init(\tau(\varphi_n)). \tag{21}$$

    Thus $M_1$ satisfies the right side of (18).

  - $M_2$ does not satisfy (20) but satisfies the following formula:

$$init(F) \wedge \neg[init(\phi_1) \vee \cdots \vee init(\phi_m) \vee succ(\phi'_1) \vee \cdots \vee succ(\phi'_k)].$$

    Thus $M_1$ satisfies the following formula:

$$\begin{aligned}&init(\tau(F)) \wedge \neg[init(\tau(\phi_1)) \vee \cdots \vee init(\tau(\phi_m)) \vee \\ &\qquad succ(\tau(\phi'_1)) \vee \cdots \vee succ(\tau(\phi'_k))].\end{aligned}$$

    So to show that the right side of the equivalence of (18) is satisfied by $M_1$, we need to show that $M_1 \models \neg succ(\phi)$. Recall that $\phi$ is a disjunction such that each disjunct $\alpha$ must correspond to a domain rule of the form $Causes(\alpha, \neg\tau(F))$, and that $\alpha$ is of the form $\alpha_1 \wedge G$ such that $\alpha_1$ does not mention fluents, and $G$ is a fluent atom that mentions an object that does not occur in $\tau(A)$. Note that $init(\alpha) \supset \neg init(\tau(F))$ is an axiom in $Succ_{\tau(A)}$, which is satisfied by $M_1$. Thus $M_1 \models \neg init(\alpha)$. This means that either $\alpha_1$ or $init(G)$ is false in $M_1$. If $\alpha_1$ is false, then $succ(\alpha)$ is false since $succ(\alpha_1)$ is the same as $\alpha_1$. Suppose that $init(G)$ is false in $M_1$. Notice that since there are no positive domain rules, and

that $G$ has an object not in $\tau(A)$ and $\tau(F)$, the pseudo-successor state axiom for $G$ in $Succ_{\tau(A)}$ must be of the form $succ(G) \equiv init(G) \wedge \Phi$. Therefore from $M_1 \models \neg init(G)$ we get $M_1 \models \neg succ(\alpha)$. Since $\alpha$ is any disjunct of $\phi$, we have proved that $M_1 \models \neg succ(\phi)$. Therefore $M_1 \models succ(\tau(F))$. Thus $M_2 \models succ(F)$.

□

## 7. Related Work

In planning, the most closely related work is the causal reasoning module in Wilkins's SIPE system (Wilkins, 1988). Wilkins writes (page 85, Wilkins, 1988): "Use of the STRIPS assumptions has made operators unacceptably difficult to describe in previous classical planners... One of the primary reasons for this is that all effects of an action must be explicitly stated... Deductive causal theories are one of the most important mechanisms used by SIPE to alleviate problems in operator representation caused by the STRIPS assumption." This is certainly one of the motivations for our system as well. In SIPE, domain rules have triggers, preconditions, conditions, and effects. Informally, when the triggers become true in the new situation, SIPE would then check in sequence to see if the preconditions were true in the old situation, and the conditions are true in the new situation. If all these conditions are true, it will then deduce the effects. For instance, a SIPE causal rule for $on(x, y)$ in the blocks world would look like:

```
Causal-rule: Not-on
Arguments: x, y, z;
Trigger: on(x,y);
Precondition: on(x,z);
Effects: not on(x,z);
```

In comparison, our domain rules are much simpler. For instance, our corresponding rule for the above SIPE rule is simply: $Causes(on(x, y) \wedge y \neq z, on(x, z))$. We do not need procedural directives like triggers. To a large degree, we can see our system as a rational reconstruction of the causal reasoning module in SIPE. As we have shown in Theorem 1, the procedure used by our system is sound under a translation to causal theories in the situation calculus. While Wilkins also gave a translation of his causal rules to formulas in the situation calculus, he did not specify an underlying logic to reason about such formulas. In fact, as shown by Lin (1995), such translations would not work.

For those familiar with PDDL, the original version by McDermott and the AIPS-98 Planning Competition Committee allows domain axioms over stratified theories. According to the manual of PDDL 1.2 (McDermott, 1998), "axioms are logical formulas that assert relationships among propositions that hold within a situation." The format for writing axioms in PDDL is as follows:

```
(:axiom
  :vars (?x ?y ...)
  :context W
  :implies P)
```

where $W$ is a formula and $P$ a literal. Axioms are treated directionally, from $W$ to $P$. The following is the rule and intention for using the axioms according to the manual:

> "The rule is that action definitions are not allowed to have effects that mention predicates that occur in the `:implies` field of an axiom. The intention is that action definitions mention 'primitive' predicates like *on*, and that all changes in truth value of 'derived' predicates like *above* occur through axioms. Without axioms, the action definitions will have to describe changes in all predicates that might be affected by an action, which leads to a complex software engineering (or 'domain engineering') problem."

It is clear from this quotation that axioms in PDDL are intended for defining "derived" predicates. They are similar to our complex fluent definitions. New versions of PDDL have extended the original version by allowing actions with durations and continuous changes. They have not considered using axioms to derive changes to "primitive" predicates like what we have done here with domain rules.

Our action domain description language, while having a very different syntax that is strongly influenced by Prolog syntax, shares much of the same ideas behind action languages (Gelfond & Lifschitz, 1999). However, unlike action languages, ours does not provide facilities for expressing the truth value of a fluent atom in a particular situation like the initial situation. Rather, it is aimed at specifying the generic effects of actions. On the other hand, it has facilities for specifying types and static relations. Most importantly, to date, action languages are either implemented directly or mapped to a nonmonotonic logic programming system rather than by compilation into a monotonic system where action effects are given explicitly, as is done here. For instance, a new SAT-based planning method would have to be implemented (e.g. McCain & Turner, 1998) for action languages. In comparison, once an action domain description is compiled to a STRIPS-like description, existing planning systems such as Blackbox (Selman & Kautz, 1999) or System R (Lin, 2001b) can be directly called.

## 8. Concluding Remarks

We have described a system for generating the effects of actions from direct action effect axioms and domain rules, among other things. We have shown the soundness of the procedure used by the system and tested it successfully in many benchmark action domains used by current AI planners. For future work, we are considering how to generalize the simple action theories in Section 6 to include context-dependent action domain descriptions like the monkey and bananas domain.

## Acknowledgments

An extended abstract of part of this paper appeared in *Proceedings of AAAI-2000.* I would like to thank the anonymous reviewers for both JAIR and AAAI'2000 as well as the associate editor in charge of this paper for JAIR for their insightful comments on earlier versions of this paper. This work was supported in part by the Research Grants Council of Hong Kong under Competitive Earmarked Research Grant HKUST6061/00E.